\pdfoutput=1

\documentclass[11pt]{article}

\usepackage[final]{acl}

\usepackage{times}
\usepackage{latexsym}

\usepackage[T1]{fontenc}

\usepackage[utf8]{inputenc}

\usepackage{microtype}

\usepackage{inconsolata}

\usepackage{graphicx}
\usepackage{hyperref}       
\usepackage{url}            
\usepackage{booktabs}       
\usepackage{amsfonts}       
\usepackage{nicefrac}       
\usepackage{microtype}      
\usepackage{xcolor}         
\usepackage{subcaption}
\usepackage{multirow}
\usepackage{algorithm}
\usepackage{algorithmic}
\usepackage{colortbl}
\usepackage{xspace}
\usepackage{listings}
\usepackage{amsmath}
\usepackage{amssymb}
\usepackage{enumitem}

\usepackage{tabularx}
\usepackage{tabulary}

\definecolor{codecomment}{HTML}{0393C4}
\definecolor{codekeyword}{HTML}{DC307F}
\definecolor{backcolor}{HTML}{F2F2F2}

\definecolor{figuremagenta}{HTML}{DC267F}

\lstdefinestyle{mystyle}{
    commentstyle=\color{codecomment},
    keywordstyle=\color{codekeyword},
    basicstyle=\ttfamily\tiny,
    breakatwhitespace=false,         
    breaklines=false,                 
    captionpos=b,                    
    keepspaces=true,                 
    showspaces=false,                
    showstringspaces=false,
    showtabs=false,                  
    tabsize=2,
    framexleftmargin=5pt,
    framextopmargin=5pt,
    framexbottommargin=8pt, 
    frame=tb, framerule=0pt,
}

\title{Skip-Layer Attention: Bridging Abstract and Detailed Dependencies in Transformers}

\author{%
  Qian Chen, Wen Wang, Qinglin Zhang, Siqi Zheng, Shiliang Zhang, \\ \textbf{Chong Deng, Hai Yu, Jiaqing Liu, Yukun Ma, Chong Zhang} \\
  Alibaba Group \\
  \texttt{\{tanqing.cq,w.wang\}@alibaba-inc.com} \\
  }

\begin{document}
\maketitle
\begin{abstract}
The Transformer architecture has significantly advanced deep learning, particularly in natural language processing, by effectively managing long-range dependencies. However, as the demand for understanding complex relationships grows, refining the Transformer's architecture becomes critical. This paper introduces Skip-Layer Attention (SLA) to enhance Transformer models by enabling direct attention between non-adjacent layers. This method improves the model's ability to capture dependencies between high-level abstract features and low-level details. By facilitating direct attention between these diverse feature levels, our approach overcomes the limitations of current Transformers, which often rely on suboptimal intra-layer attention. Our implementation extends the Transformer's functionality by enabling queries in a given layer to interact with keys and values from both the current layer and one preceding layer, thus enhancing the diversity of multi-head attention without additional computational burden. Extensive experiments demonstrate that our enhanced Transformer model achieves superior performance in language modeling tasks, highlighting the effectiveness of our skip-layer attention mechanism.
\end{abstract}

\section{Introduction}

The Transformer architecture has made notable strides in the field of large language models (LLMs) \citep{DBLP:conf/naacl/DevlinCLT19, Radford2018ImprovingLU, Radford2019LanguageMA, Brown2020LanguageMA, Ouyang2022TrainingLM,DBLP:journals/corr/abs-2303-08774}. These models have impressively tackled a variety of tasks, including natural language understanding \citep{DBLP:conf/iclr/HendrycksBBZMSS21}, general question answering \citep{DBLP:journals/corr/abs-2311-12022}, coding \citep{DBLP:journals/corr/abs-2107-03374}, mathematics \citep{DBLP:journals/corr/abs-2110-14168}, and scientific knowledge \citep{DBLP:conf/emnlp/ChenYKLWMXWX23}. However, as data grows more complex and relationships more intricate \citep{DBLP:journals/corr/abs-2311-12022}, there's a need for ongoing improvements in the architecture to keep up with these challenges.

The primary strength of the Transformer lies in its self-attention mechanism, which allows each element in the input sequence to compare directly with every other element, thereby capturing dependencies regardless of their distance \citep{DBLP:conf/nips/VaswaniSPUJGKP17}. Nevertheless, this design faces limitations when handling more complex relationships. The original intra-layer attention in Transformers is often inadequate for capturing the deeper interactions (i.e., high-level abstract features and low-level details) demanded by more complex tasks \citep{DBLP:conf/atal/Tenenbaum18,DBLP:conf/naacl/YangYDHSH16}.

\begin{figure*}
  \centering
  \includegraphics[width=0.75\textwidth]{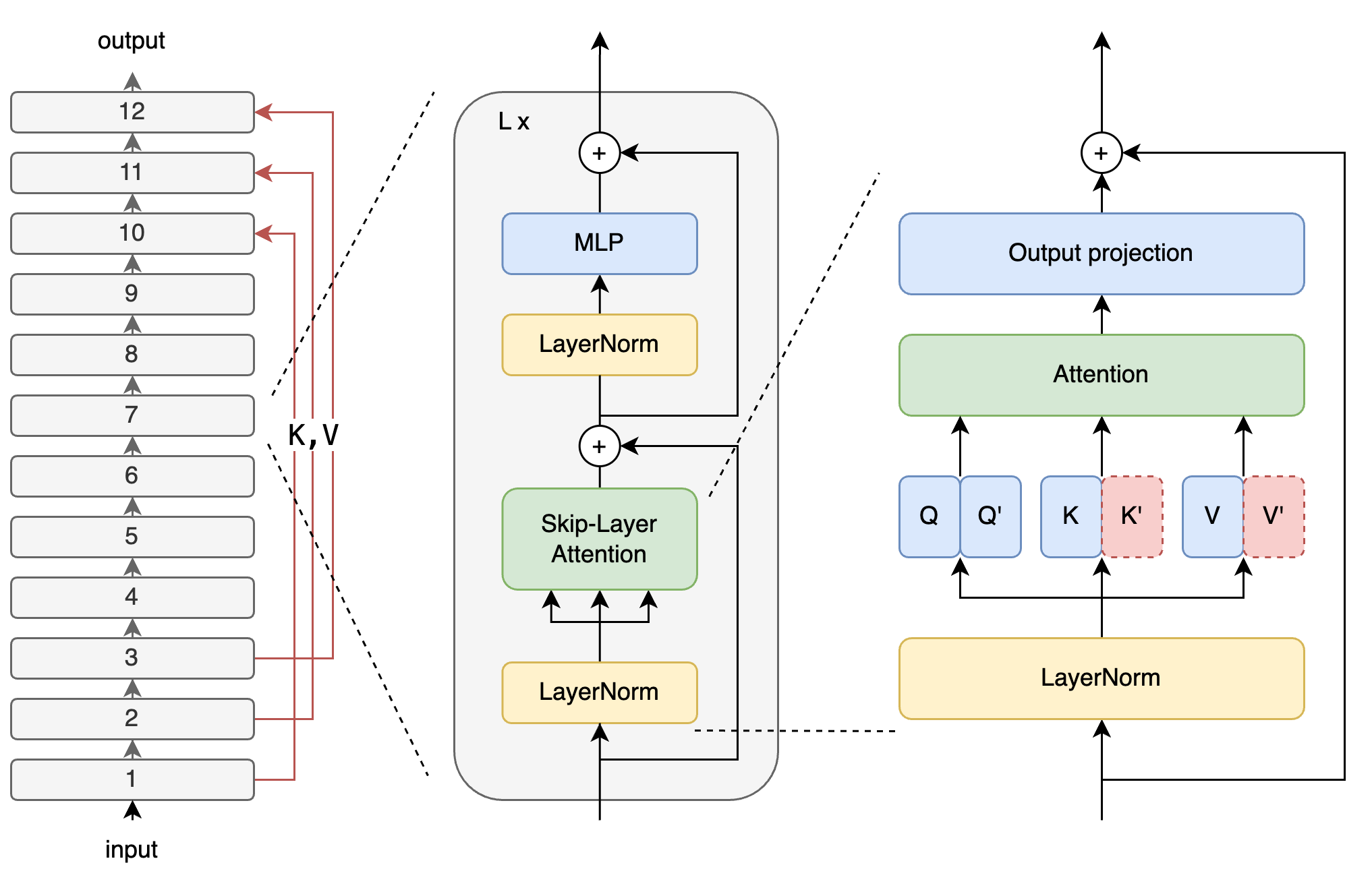}
  \caption{Model architecture of the Transformer with skip-layer attention. The left figure illustrates a Transformer model with 12 layers, each equipped with an additional skip-layer attention connection (e.g., layer 1 to layer 10, layer 2 to layer 11, layer 3 to layer 12). The center figure provides a zoomed-in view of each layer, highlighting the skip-layer attention and MLP sublayers. The right figure details the skip-layer attention mechanism, with red indicating keys and values from the preceding layer.}
  \label{fig:model}
\end{figure*}

To address these limitations, researchers have explored various methods employed in earlier models such as ResNet \citep{DBLP:conf/cvpr/HeZRS16} and Highway Networks \citep{DBLP:journals/corr/SrivastavaGS15}. Our goal is to refine inter-layer interactions within Transformers. Drawing inspiration from DenseNet \citep{DBLP:conf/cvpr/HuangLMW17} in convolutional neural networks (CNNs), which employs dense cross-layer connections to facilitate feature propagation, we propose a novel Skip-Layer Attention (SLA) approach to enhance the Transformer model. Our implementation augments the Transformer's capabilities by permitting queries in a given layer to interact not only with keys and values from the current layer but also from the preceding layer. This method enriches the diversity of multi-head attention, while maintaining the same computational efficiency. Unlike DenseNets, which focus on identical tokens across layers, our strategy connects both identical and distinct tokens, thereby enhancing the model's capacity to capture and incorporate both abstract and detailed dependencies.
Our contributions are as follows:
\begin{itemize}
\item We propose a novel mechanism that enables direct attention between non-adjacent layers, enhancing the ability to capture dependencies between high-level abstract features and low-level details.
\item Our method extends the Transformer's functionality without significantly increasing computational complexity, making it practical for large-scale applications.
\item Through extensive experiments against Transformer baselines, we demonstrate the effectiveness of our enhanced architecture in language modeling tasks.
\end{itemize}

\section{Related Work}
The concept of enhancing network connectivity originates from earlier architectures such as ResNet \citep{DBLP:conf/cvpr/HeZRS16}, which introduces residual connections. These residual connections enable the training of much deeper networks by facilitating the flow of gradients during backpropagation. Highway Networks \citep{DBLP:journals/corr/SrivastavaGS15} introduce gated connections to regulate information flow across layers, making the end-to-end training of deep networks more feasible.

DenseNet \citep{DBLP:conf/cvpr/HuangLMW17} advances this idea by creating an intricate connectivity pattern where each layer connects to every other layer in a feed-forward manner. This dense connectivity promotes feature reuse and significantly reduces the number of parameters, directly inspiring the skip-layer connectivity pattern explored in our work. However, DenseNet primarily targets CNNs and is mainly applied to computer vision tasks, with connections occurring between the same tokens in subsequent layers. Our approach extends this concept to Transformers by incorporating direct connections among both identical and distinct tokens.

More recently, \citet{brandon2024reducing} propose sharing key/value heads across adjacent layers to reduce the size of the KV cache. This strategy draws inspiration from the success of Multi-Query Attention \citep{DBLP:journals/corr/abs-1911-02150} and Grouped-Query Attention \citep{DBLP:conf/emnlp/AinslieLJZLS23}. Our skip-layer attention, however, enables direct attention between non-adjacent layers, thus bridging the dependencies between high-level abstract features and low-level details. This approach encapsulates the depth of interactions required by more demanding tasks.

\section{Method}
Our novel Transformer model enhances the standard Transformer architecture by incorporating skip-layer attention, aimed at improving the information flow between non-adjacent layers. As shown in Figure \ref{fig:model}, our approach replaces the conventional multi-head attention sublayer with skip-layer attention sublayer. This upgrade establishes direct connections between layers that are not immediately adjacent, thereby promoting a more efficient and comprehensive exchange of information across the entire network.

Our model retains the Transformer's original multi-head attention mechanism, utilizing queries, keys, and values, but extends this framework by enabling queries in a given layer to interact with keys and values from both the current layer and one preceding layer. This extension is fundamental to the implementation of skip-layer attention.

We denote the queries after linear projection for all heads as the tensor \( Q \in \mathbb{R}^{h \times T \times d} \), where \( h \) is the number of heads, \( T \) is the sequence length, and \( d \) is the hidden size for each head. Similarly, the keys and values are represented as tensors \( K \in \mathbb{R}^{h \times T \times d} \) and \( V \in \mathbb{R}^{h \times T \times d} \), respectively. The number of skip layers is denoted as \( n_l \), and the number of skip heads as \( n_h \).

The attention mechanism for each head is described by the following equations:

\begin{equation*}
\begin{aligned}
\small
\text{H}_i^l =
    \begin{cases} 
    \text{Att}(Q^l_i, K^l_i, V^l_i) & \text{if } i \in \{1, ..., h - n_l\} \\
    \text{Att}(Q^l_i, K^{l-n_l}_i, V^{l-n_l}_i) & \text{if } i \in \{h - n_l + 1, ..., h\}
    \end{cases}
\end{aligned}
\end{equation*}

\noindent where $i$ represents the index of the head and $l$ represents the index of the layer. This formulation allows our model to effectively bridge abstract and detailed dependencies and improve information flow throughout the network. The core implementation in PyTorch can be found in Appendix \ref{sec:appendix_code}.

\section{Experimental Setup}
\subsection{Dataset}
We use the OpenWebText corpus\footnote{\url{https://skylion007.github.io/OpenWebTextCorpus/}}, an open-source recreation of the WebText dataset. It comprises approximately 8 million documents sourced from Reddit-linked web content. The corpus is divided into a training set with about 9 billion tokens and a validation set with around 4 million tokens.

\subsection{Training Setup}
Training batches contain 524,288 tokens, stabilizing the process across model scales. The training spans a maximum of 18,000 steps, processing roughly 9.4 billion tokens, equivalent to one full epoch on the training set. We experiment with three sizes of the GPT-2 architecture: GPT2 (124M), GPT2-Medium (350M) and GPT2-Large (774M). Flash Attention \citep{DBLP:journals/corr/abs-2307-08691} is incorporated to accelerate attention operations. Training is conducted on NVIDIA V100 GPUs with 32GB of memory. We experiment with sequence lengths of 4096, 8192, and 16,384 tokens to explore long-range dependencies. Code implementation and optimization are managed using the nanoGPT framework\footnote{\url{https://github.com/karpathy/nanoGPT/}}.

All models start with a learning rate of 1.5e-4, determined to offer a good balance between rapid convergence and stability. Model performance is evaluated based on the loss on the validation dataset.

\begin{table}
  \caption{Optimal number of skip layers.}
  \label{tab:search_layer}
  \centering
  \scalebox{0.9}{
  \begin{tabular}{llcc}
    \toprule
    \#SkipLayer    & \#SkipHead     & Loss & Abs. Impr. \\
    \midrule
    0 (Baseline) & 0 & 3.3826 & - \\
    1 & 6 & 3.4030 & -0.0204 \\
    3 & 6 & 3.2958 & 0.0868 \\
    6 & 6 & 3.2853 & 0.0973 \\
    \textbf{9} & \textbf{6} & \textbf{3.2750} & \textbf{0.1076} \\
    11 & 6 & 3.3734 & 0.0092 \\
    \bottomrule
  \end{tabular}
  }
\end{table}

\begin{table}
 \caption{Optimal number of skip heads.}
 \label{tab:search_head}
  \centering
  \scalebox{0.9}{
  \begin{tabular}{llcc}
    \toprule
\#SkipLayer &	\#SkipHead	&Loss	& Abs. Impr. \\
\midrule
0 (Baseline) & 0 & 3.3826 & - \\
9 & 3 & 3.3543 & 0.0283 \\
9 & 6 & 3.2750 & 0.1076 \\
\textbf{9} & \textbf{9} & \textbf{3.2497} & \textbf{0.1329} \\
9 & 12 & 3.2643 & 0.1183 \\
    \bottomrule
  \end{tabular}
  }
\end{table}

\begin{table*}
  \caption{Model size and sequence length variations.}
  \label{tab:size_length}
  \centering
  \begin{tabular}{llcccc}
    \toprule
Model & Length & Baseline & Skip-Layer Attention & Abs. Impr. & Training Speedup  \\
\midrule
\multirow{3}{*}{GPT2(124M)} & 4096 & 3.1858 & 3.1762 & 0.0096 & -1.69\%\\
 & 8192 & 3.2077 & 3.2020 & 0.0057 & -0.28\%\\
 & \textbf{16384} & \textbf{3.3826} & \textbf{3.2497} & \textbf{0.1329} & \textbf{0.83\%} \\
 \midrule
\multirow{2}{*}{GPT2-Medium(350M)} & 4096 & 2.9538 & 2.9399 & 0.0139 & -0.76\% \\
 & \textbf{8192} & \textbf{3.0335} & \textbf{2.9506} & \textbf{0.0829} & \textbf{-2.34\%} \\
 \midrule
GPT2-Large(774M) & 4096 & 2.8271 & 2.8156 & 0.0115 & -1.24\% \\
    \bottomrule
  \end{tabular}
\end{table*}

\section{Result}
\subsection{Number of skip layers}
We initially explore the impact of varying the number of skip layers using a GPT-2 model (124M parameters) as our default backbone. This model has a hidden size of 768, 12 heads, 12 layers, and supports a sequence length of 16,384. The default number of skip heads is set to 6. As shown in Table \ref{tab:search_layer}, our findings indicate that the optimal performance enhancement via the skip-layer attention method is achieved with 9 skip layers, resulting in a substantial absolute improvement of 0.1076 over the baseline. Configurations with 3 and 6 skip layers also demonstrate notable progress. However, employing just a single skip layer yields no benefit; this might be attributed to the similarity between the key and value heads among adjacent layers, as discussed in \citet{brandon2024reducing}. Similarly, setting the number of skip layers to 11 does not produce noticeable advancements, possibly due to the presence of only one skip-layer attention in this setup. Based on these results, we recommend that 3/4 of the total number of layers is the most effective number of skip layers.

\subsection{Number of skip heads}
Subsequently, we investigate the impact of varying the number of skip heads, also using a GPT-2 model (124M) and a sequence length of 16,384 as our default backbone. The default number of skip layers is set to 9, based on the optimal configuration identified in the previous section. As presented in Table \ref{tab:search_head}, our results indicate that the optimal performance enhancement is observed with 9 skip heads, yielding a significant absolute improvement of 0.1329 over the baseline. Configurations with 6 and 12 skip heads also demonstrate commendable improvements. Notably, the configuration with 12 skip heads suggests that the keys and values from the last 3 layers all use the keys and values from the first 3 layers. This implies that direct attention modeling between high-level abstract features and low-level detail features is more important than attention modeling purely between high-level abstract features. Conversely, employing only 3 skip heads shows no substantial benefit, indicating that more skip-attention heads are necessary than the original attention heads. Based on these findings, we recommend setting the number of skip heads to 3/4 of the total number of heads.

\subsection{Model Size and Sequence Length Variations}
In this section, we explore the effects of varying model sizes and sequence lengths on performance. We maintain the default configuration of using 3/4 of the total number of layers as skip layers and 3/4 of the total number of heads as skip heads. Due to the 32GB memory limit of the V100 GPU, we restrict our tests to model sizes and sequence lengths that do not trigger out-of-memory (OOM) errors when using Distributed Data Parallel (DDP).

As shown in Table \ref{tab:size_length}, we observe an absolute improvement of 0.1329 over the baseline when using GPT-2 (124M) with a sequence length of 16,384. However, no significant improvements are noted for sequence lengths of 4,096 and 8,192. This suggests that longer sequences benefit more from our skip-layer attention method, likely because they encompass more abstract and detailed dependencies.
Furthermore, for a sequence length of 8,192, GPT-2 Medium (350M) achieves an absolute improvement of 0.0829 over the baseline, while GPT-2 (124M) shows no noticeable improvement. This indicates that larger models gain a greater advantage from our skip-layer attention method. 

The training time with our skip-layer attention method is slightly longer than the baseline, experiencing a maximum decrease in training speed of 2.34\%. This is likely due to the additional storage requirements for keys and values in the lower layers.

\section{Conclusion}
In this paper, we propose a Skip-Layer Attention (SLA) mechanism to enhance the Transformer architecture's ability to capture complex dependencies within input data. By enabling direct attention between non-adjacent layers, our approach improves the model's capacity to integrate high-level abstract features with low-level details without significantly increasing computational complexity. Extensive experiments demonstrate that our enhanced Transformer model outperforms standard Transformer baselines in language modeling tasks, validating the effectiveness of our method. Our findings pave the way for further innovations in optimizing neural network architectures.

\section{Limitations}
While our research has demonstrated the effectiveness of skip-layer attention in Transformer models, several avenues for future work remain to be explored:

\textbf{Scaling to Larger Models}: Future research could extend skip-layer attention to larger Transformer architectures, such as GPT-3 \citep{Brown2020LanguageMA} and beyond, to assess its effectiveness across different scales and complexity levels. Evaluating the performance and efficiency on these larger models will offer valuable insights into its scalability.

\textbf{Real-World Applications}: Evaluating the skip-layer attention mechanism in various real-world applications, such as natural language understanding \citep{DBLP:conf/iclr/HendrycksBBZMSS21}, general question answering \citep{DBLP:journals/corr/abs-2311-12022}, coding \citep{DBLP:journals/corr/abs-2107-03374}, and mathematics \citep{DBLP:journals/corr/abs-2110-14168}, will be critical to fully understand its practical benefits and limitations.

\textbf{Beyond Text}: Extending the applicability of the skip-layer attention mechanism to other domains, such as computer vision and speech processing, will help determine its versatility and potential for cross-modal impact.

\textbf{Ablation Studies}: Conducting comprehensive ablation studies to understand the contributions of different components within the skip-layer attention mechanism could provide deeper insights. For instance, exploring the impact of connections to multiple preceding layers rather than just one could reveal additional enhancements and inform design choices.


\bibliography{custom}

\clearpage
\appendix
\section{Appendix}
\label{sec:appendix}

\subsection{Code}
\label{sec:appendix_code}
In this paper, the model code is based on nanoGPT, which is publicly available on GitHub\footnote{\url{https://github.com/karpathy/nanoGPT/}}. The implementation of the skip-layer attention method is detailed in the code listing provided below.

\begin{center}
\tiny
\begin{minipage}[t]{2\linewidth}%
\lstset{backgroundcolor=\color{backcolor}}

\begin{lstlisting}[language=Python,mathescape=true]
import torch.nn.functional as F

class CausalSelfSkipLayerAttention(nn.Module):

    def __init__(self, config):
        super().__init__()
        assert config.n_embd % config.n_head == 0
        # key, query, value projections for all heads, but in a batch
        self.c_attn = nn.Linear(config.n_embd, 3 * config.n_embd, bias=config.bias)
        # output projection
        self.c_proj = nn.Linear(config.n_embd, config.n_embd, bias=config.bias)
        # regularization
        self.attn_dropout = nn.Dropout(config.dropout)
        self.resid_dropout = nn.Dropout(config.dropout)
        self.n_head = config.n_head
        self.n_embd = config.n_embd
        self.dropout = config.dropout
        self.num_skip_layer = config.num_skip_layer
        self.split_num = config.n_head - config.num_skip_head

    def forward(self, x, prev_k_list=[], prev_v_list=[]):
        B, T, C = x.size()  # batch size, sequence length, n_embd
        
        q, k, v = self.c_attn(x).split(self.n_embd, dim=2)
        k = k.view(B, T, self.n_head, C // self.n_head).transpose(1, 2)  # (B, nh, T, hs)
        q = q.view(B, T, self.n_head, C // self.n_head).transpose(1, 2)  # (B, nh, T, hs)
        v = v.view(B, T, self.n_head, C // self.n_head).transpose(1, 2)  # (B, nh, T, hs)

        # causal self-attention; Self-attend: (B, nh, T, hs) x (B, nh, hs, T) -> (B, nh, T, T)
        # efficient attention using Flash Attention CUDA kernels
        if len(prev_k_list) >= self.num_skip_layer and len(prev_v_list) >= self.num_skip_layer:
            k_combine = torch.cat([k[:, :self.split_num, :, :], prev_k_list[-self.num_skip_layer]], dim=1)
            v_combine = torch.cat([v[:, :self.split_num, :, :], prev_v_list[-self.num_skip_layer]], dim=1)
            y = F.scaled_dot_product_attention(q, k_combine, v_combine, attn_mask=None, 
                                               dropout_p=self.dropout if self.training else 0, 
                                               is_causal=True)
        else:
            y = F.scaled_dot_product_attention(q, k, v, attn_mask=None, 
                                               dropout_p=self.dropout if self.training else 0, 
                                               is_causal=True)

        y = y.transpose(1, 2).contiguous().view(B, T, C)
        # output projection
        y = self.resid_dropout(self.c_proj(y))
        return y, k[:, self.split_num:, :, :], v[:, self.split_num:, :, :]
\end{lstlisting}
\end{minipage}
\end{center}

\end{document}